\def\year{2021}\relax
\documentclass[letterpaper]{article} 
\usepackage{aaai21}  
\usepackage{times}  
\usepackage{helvet} 
\usepackage{courier}  
\usepackage[hyphens]{url}  
\usepackage{graphicx} 
\usepackage{natbib}  
\usepackage{caption} 

\usepackage{algorithm}
\usepackage{algorithmicx}
\usepackage{algpseudocode}
\usepackage{amsmath}
\usepackage{booktabs}

\usepackage{threeparttable}

\usepackage[scriptsize]{subfigure}
\usepackage{pgfplots}
\usepackage{filecontents}
\usepackage{setspace}
\usepackage{multirow}
\usepackage{listings}
\usepackage{amssymb}
\usepackage{enumitem,amssymb}

\usepackage{pifont}
\newcommand{\cmark}{\ding{51}}%
\newcommand{\xmark}{\ding{55}}%
\usepackage{xcolor}

\title{Improving Tree-Structured Decoder Training for Code Generation \\
	via Mutual Learning}
\author{
    Binbin Xie\textsuperscript{\rm 1,2}{\rm{,}\ } 
    Jinsong Su\textsuperscript{\rm 1,2 \footnote{Corresponding author.}}{\rm{,}\ } 
    Yubin Ge\textsuperscript{\rm 3}{\rm{,}\ } 
    Xiang Li\textsuperscript{\rm 4}{\rm{,}\ } 
    Jianwei Cui\textsuperscript{\rm 4}{\rm{,}\ } \\
    Junfeng Yao\textsuperscript{\rm 1} {\rm{and}\ } 
    Bin Wang\textsuperscript{\rm 4}
}

\affiliations{

   1. Xiamen University, Xiamen, China \\
   2. Peng Cheng Laboratory, Shenzhen, China \\
   3. University of Illinois at Urbana-Champaign, Urbana, IL 61801, USA \\
   4. Xiaomi AI Lab, Beijing, China \\
   xdblb@stu.xmu.edu.cn, \ \ \{jssu, yao0010\}@xmu.edu.cn, \\
    yubinge2@illinois.edu, \ \
   \{lixiang21, cuijianwei, wangbin11\}@xiaomi.com,

}

\begin{document}

\maketitle

\begin{abstract}
Code generation aims to automatically generate a piece of code given an input natural language utterance.
Currently,
among dominant models, it is treated as a sequence-to-tree task,
where a decoder outputs a sequence of actions corresponding to the pre-order traversal of an Abstract Syntax Tree.
However,
such a decoder only exploits the pre-order traversal based preceding actions,
which are insufficient to ensure correct action predictions.
In this paper,
we first throughly analyze the context modeling difference between neural code generation models with different traversals based decodings (preorder traversal vs breadth-first traversal),
and then propose to introduce a mutual learning framework to jointly train these models.
Under this framework,
we continuously enhance both two models via mutual distillation, which involves
synchronous executions of two one-to-one knowledge transfers at each
training step.
More specifically, we alternately choose one model as the student and the other as its teacher, and require the student to fit the
training data and the action prediction distributions of its teacher.
By doing so,
both models can fully absorb the knowledge from each other and thus could be improved simultaneously.
Experimental results and in-depth analysis on several benchmark datasets demonstrate the effectiveness of our approach.
We release our code at \url{https://github.com/DeepLearnXMU/CGML}.
\end{abstract}

\section{ Introduction}
As an indispensable text generation task,
code generation mainly focuses on automatically generating a snippet of code given a natural language (NL) utterance.
Due to its great potential in facilitating software development and revolutionizing end-user programming,
it has always been one of hot research topics in the communities of natural language processing and software engineering.

To achieve this goal, previous studies for code generation either require manually-designed grammar and lexicons \cite{DBLP:phd/ndltd/Zettlemoyer09,DBLP:conf/emnlp/ZettlemoyerC07} or exploit features for candidate logical forms ranking \cite{DBLP:conf/acl/LiangJK11}.
With the prosperity of deep learning in natural language processing,
dominant models have now evolved into neural models,
which possess the powerful capacity of feature learning and representation.
Among them,
the most prevalent one is the sequence-to-tree (Seq2Tree) models \cite{DBLP:conf/acl/DongL16,DBLP:conf/acl/YinN17,DBLP:conf/acl/RabinovichSK17,DBLP:conf/acl/LapataD18,DBLP:conf/nips/ShinABP19,DBLP:conf/aaai/SunZXSMZ20}, all of which are based on an encoder-decoder framework.
Specifically,
an encoder learns the word-level semantic representations of an input NL utterance,
and then a tree-structured decoder outputs a sequence of actions, which corresponds to the pre-order traversal of an Abstract Syntax Tree (AST) and can be further converted into the code,
1) the structure of AST can be used to shrink the search
space, ensuring the generation of well-formed code;
and
2) the structural information of AST helps to model the information flow within the neural network, which naturally reflects the recursive structure of programming languages.

Despite the success of the above models,
limited by the decoding manner based on the pre-order traversal,
there still exists a serious defect in their decoders.
Specifically, at each timestep, the current action can be only predicted from the vertical preceding actions.
Since the dependencies of actions may come from other directions, i.e. the actions in
horizontal direction,
it may be insufficient to accurately predict the current action only with preceding actions in vertical direction.
This happens especially when an important action locates in horizontal direction.
Another possible case would be the strength of the dependencies on preceding actions turn to be weak with the increasing distances to those actions.
Moreover,
prediction errors in preceding actions sometimes even hurt the prediction of current action.
Therefore,
it is worth exploring the context in other direction to complement the currect context of the pre-order traversal based decoder.

In this paper,
we first explore the Seq2Tree model with breadth-first traversal based decoding,
and then analyze the
context modeling difference between different traversals
based decodings (pre-order traversal vs. breath-first traversal).
More importantly,
we propose to introduce a mutual learning framework \cite{DBLP:conf/cvpr/ZhangXHL18} to jointly train two Seq2Tree models with different traversals based decodings,
both of which are expected to benefit each other.
Our motivation stems from the fact that these two models are able to capture the context from different directions.
That is,
the conventional Seq2Tree model focuses on the pre-order traversal based preceding actions,
while the other one mainly pays attention to the preceding actions based on breadth-first traversal.
Hence, such difference brings the potential in enhancing both models.

Under the proposed framework,
we continuously improve both models via mutual distillation,
which involves synchronous executions of two one-to-one transfers at each training step.
More specifically,
we alternately choose one model as the student and the other as its teacher,
and require the student to fit the training data and match action prediction distributions of its teacher.
By doing so,
both models can fully absorb the knowledge from each other and thus could be improved simultaneously.

The contributions of our work are summarized as follows:
\begin{itemize}
	\item We explore the neural code generation model with breadth-first traversal based decoding, and point out that models with different traversals can be complementary to each other in context modeling.
	\item We propose to introduce a mutual learning based model training framework for code generation, where Seq2Tree models with different traversals based decodings can continuously enhance each other.
	\item On several commonly-used datasets, we demonstrate the effectiveness and generality of our framework.
\end{itemize}

\section{Background}
\label{background}
In this work,
we choose \textsc{Tranx} \cite{DBLP:conf/emnlp/YinN18} as our basic model,
which has been widely used due to its competitive performance
\cite{DBLP:conf/acl/YinN19,DBLP:conf/nips/ShinABP19,DBLP:conf/acl/XuJYVN20}.
Note that both our explored decoder with breadth-first traversal and the mutual learning based model training framework are also applicable to other Seq2Tree models.
In the following subsections,
we first introduce how to convert NLs to code using \textsc{Tranx},
and then give a detailed description of the architecture of \textsc{Tranx}.

\subsection{Converting NLs to Code using TRANX}
\begin{figure*}[ht]
	\centering
	\includegraphics[width=0.84\textwidth]{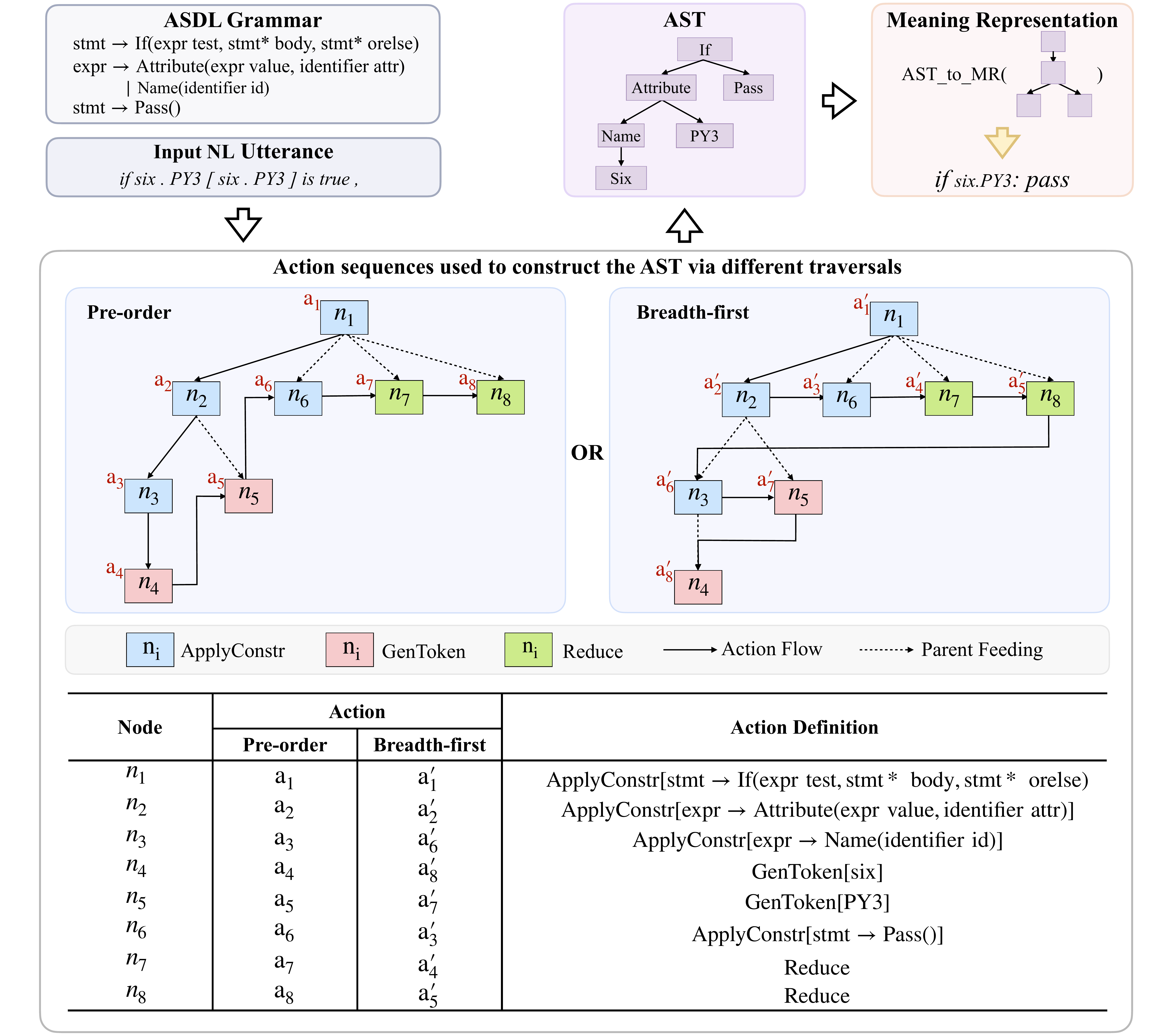}
	\caption{The procedure of converting an input NL utterance into a snippet of code.}
	\label{figure1} 
\end{figure*}

Typically,
\textsc{Tranx} introduces ASTs as the intermediate meaning representations (MRs).
Figure \ref{figure1} gives an example of the NL utterance-to-code conversion for the Python code ``\textit{if six.PY3: pass}''.
Concretely,
such a conversion involves two stages:
(1) the transition from the NL utterance into an AST.
During this process,
Tranx outputs a sequence of ASDL formalism based actions,
corresponding to the AST in pre-order traversal;
and
(2) A user specified function $\textrm{AST\_{to}\_{MR}(*)}$ is called to convert the generated \ AST into the code.

Specifically,
at each timestep,
one of three types of actions is evoked to expand the non-terminal node of the partial AST:

(1) \textsc{ApplyConstr[$c$]} actions apply a constructor $c$ to the opening
composite frontier field which has the same type as
$c$, populating the opening node using the fields in $c$.
If the frontier field has sequential cardinality, the action appends
the constructor to the list of constructors held by the field.
Back to Figure \ref{figure1},
“$n_1$:\textsc{ApplyConstr}[stmt $\to$ If(expr test, stmt* body, stmt* orelse)]” acts on the root of the AST and has three fields to be expended.

(2) \textsc{Reduce} actions mark the completion of the generation of child values for a field with optional (?) or multiple ($*$) cardinalities.
In Figure \ref{figure1},
“$n_7$:\textsc{Reduce}” means the completion of the field “stmt* body” of
“$n_1$:\textsc{ApplyConstr} [stmt $\to$ If(expr test, stmt* body, stmt* orelse)]”.

(3) \textsc{GenToken[$v$]} actions populate a (empty) primitive frontier
field with a token $v$.
Specifically,
the field of node with primitive types should be filled with \textsc{GenToken}[$v$] action. For instance,
“$n_4$:\textsc{GenToken}[$six$]” means the token “$six$” fills the identifier field of “\textsc{ApplyConstr}[expr $\to$ Name(identifier id)]”.

\subsection{TRANX}

{\textsc{Tranx}} is based on an attentional encoder-decoder framework,
where
a BiLSTM encoder is used to learn word-level semantic representations $\{h_i\}^{|X|}_{i=1}$ of the input NL utterance $X$,
converted by an LSTM decoder into a sequence of tree-constructing actions.

Specifically,
during the timestep $t$ of decoding, its LSTM cell reads the embedding $e(a_{t-1})$ of the previous action ${a_{t-1}}$, the temporary hidden state ${\tilde{s}_{t-1}}$,
the vector $p_t$ concatenated by the hidden state of the parent AST node and the type embedding of the current AST node,
and the previous hidden state ${s_{t-1}}$ to generate the hidden state vectors ${s_t}$ and ${\tilde{s}_{t}}$ as:
\begin{alignat}{2}
	s_t &= f_{\text{LSTM}}( [e(a_{t-1});\tilde{s}_{t-1};p_t], s_{t-1}), \label{lstm_func}\\
	\tilde{s}_t &= \tanh(W_s[c_t;s_t]),
\end{alignat}
where $W_s$ is a parameter matrix and the attentional context vector $c_t$ is produced from $\{h_i\}$.

Finally,
the probability of applying the action $a_t$ is computed according to the type of the current node specified by the corresponding action of its parent node:

(1) \textbf{\textit{Primitive} type.}
In this case,
\textsc{Tranx} takes an action $a_t$=$\textsc{GenToken}[\emph{v}]$,
where the token $v$ can be either generated directly or copied from the input NL utterance as follows:
\begin{equation}
	\begin{aligned}
		\begin{split}
			&p(a_t=\textsc{GenToken}[\emph{v}] |a_{<t}, \mathbf{x}) \\
			=\ &p(gen|a_{<t}, \mathbf{x})\cdot p_{gen}(\emph{v}|a_{<t}, \mathbf{x}) \\
			+\ &(1 - p(gen|a_{<t}, \mathbf{x}))\cdot p_{copy}(\emph{v}|a_{<t}, \mathbf{x}).
		\end{split}
	\end{aligned}
\end{equation}
Here, we use three softmax functions based on $\tilde{s}_t$ to the probability $p(gen|a_{<t},\mathbf{x})$ of choosing the generation operation, the probability $p_{gen}(v|a_{<t},\mathbf{x})$ of generating $v$ and the probability $p_{copy}(v=x_i|a_{<t},\mathbf{x})$ of selecting to copy $x_i$, respectively.

(2) \textbf{\emph{Composite} type.}
On this condition, \textsc{Tranx} either applies an \textsc{ApplyConstr} action to produce the current node,
or employs a \textsc{Reduce} operation to terminate some node expansion from the AST parent node.
Formally, the probability $p(a_t|a_{<t}, \mathbf{x})$ of the action $a_t$ is computed as
$\text{softmax}(e(a_t)^\mathsf{T} W_a\tilde{s}_t)$, where $W_a$ is the parameter matrix of the linear mapping.

\subsubsection{Training Objective.}
Given a training corpus $D=\{(\mathbf{x,a})\}$, the training objective $J_\textrm{MLE}(D;\theta)$  is defined as
\begin{alignat}{2}
	J_\textrm{MLE}(D;\theta) &= \sum_{(\mathbf{x,a})\in D} J_\textrm{MLE}(\mathbf{x,a};\theta),\\
	J_\textrm{MLE}(\mathbf{x,a};\theta) &= -\frac{1}{T}\sum_{t=1}^{T} {\log{p(a_t|a_{<t},\mathbf{x};\theta)}}.
	\label{lossfunction} 
\end{alignat}

\section{Our Approach}
In this section, we first briefly describe two Seq2Tree models with different traversals (pre-order traversal and breadth-first traversal) based decodings.
Then, we propose to introduce mutual learning \cite{DBLP:conf/cvpr/ZhangXHL18} based model training framework into code generation,
where these two models can be enhanced simultaneously.

\subsection{Seq2Tree Models}
\subsubsection{\textsc{Tranx}.}
We omit the description of \textsc{Tranx}, which has been provided in Background section.

\begin{figure*}[th]\small
	\centering
	\includegraphics[width=0.90\textwidth]{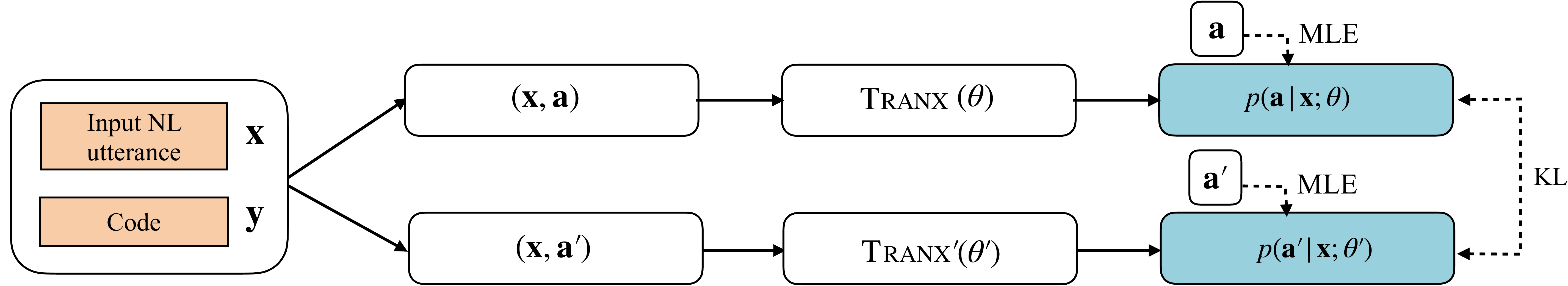}	
	\caption{Mutual learning based model training framework.
		Particularly, \textsc{Tranx} and $\textsc{Tranx}'$ share the same encoder.} 
	\label{figure2} 
\end{figure*}

\begin{algorithm}[th]
	\small
	\caption{The training procedure of our framework.}
	\label{conjugateGradient}
	\begin{algorithmic}[1]
		\Require
		Training set $D$,
		validation set $D_v$		
		\Repeat
		\Repeat
		\State Load a batch size of instances $B \in D$
		\State $J(B;\theta) = 0$
		\State $J(B;\theta')=0$
		\For{ each instance $(\mathbf{x},\mathbf{a}) \in B$}\label{line7}
		\State Traverse the AST of $\mathbf{a}$ in a breadth-first order to form a reference action sequence $\mathbf{a}'$ for $\textsc{Tranx}'$
		\State Calculate the loss $J(\mathbf{x,a};\theta)$ of \textsc{Tranx}
		\State Calculate the loss $J(\mathbf{x,a'};\theta')$ of $\textsc{Tranx}'$
		\State $J(B;\theta)$ += $J(\mathbf{x,a};\theta)$
		\State $J(B;\theta')$ += $J(\mathbf{x,a'};\theta')$
		\EndFor
		\State Minimize $J(B,\theta)$ to update $\theta$
		\State Minimize $J(B,\theta')$ to update $\theta'$
		\State Save the best models according to their performance on $D_v$ \label{line15}
		\Until no more batches
		\Until both $\theta$ and $\theta'$ converge
	\end{algorithmic}
\end{algorithm}

\subsubsection{$\textsc{Tranx}'$.}
As a variant of $\textsc{Tranx}$, $\textsc{Tranx}'$ provides a new perspective of code generation,
where the AST is generated via breadth-first traversal and thus the AST context in horizontal direction can be leveraged for action predictions.

To train $\textsc{Tranx}'$,
we traverse the AST of each training example in breadth-first order to generate an reference action sequence $\mathbf{a}'$.
As shown in the middle of Figure \ref{figure1},
for the code ``\textit{if six.PY3 : pass}'',
we traverse its AST in the breadth-first order and then reorganize the corresponding actions into an action sequence: $\mathbf{a'}$=$\text{a}'_1$, $\text{a}'_2$,..., $\text{a}'_8$.
Note that although reference action sequences of $\textsc{Tranx}$ and $\textsc{Tranx}'$ are different,
they are from the same AST.

Likewise,
we then parameterize the prediction probability using a LSTM netowrk:
$p(\mathbf{a'}|\mathbf{x}) = \prod_{t'=1}^{T}p(a'_{t'}|a'_{<t'},\mathbf{x})$.
The main equations of this LSTM are almost the same as those of \textsc{Tranx},
the only difference is that the definitions of preceding actions are different.
Very importantly,
$e(a_{t-1})$ of Equation \ref{lstm_func} is replaced with $e(a'_{t'-1})$.

Apparently,
during different traversals based decodings,
\textsc{Tranx} and $\textsc{Tranx}'$ focus on the AST-based context from different directions.
For example,
for the action prediction of the considered node $n_6$,
\textsc{Tranx} exploits the context encoded by nodes $n_1$ and $n_5$
while $\textsc{Tranx}'$ makes use of the context of its sibling $n_2$.
Intuitively,
both nodes $n_2$ and $n_1$ are closely-related nodes of $n_6$,
possessing important impact on the action prediction of $n_6$.
Therefore,
we believe that \textsc{Tranx} and $\textsc{Tranx}'$ can complement each other.


\subsection{Mutual Learning based Model Training}

As illustrated in Figure \ref{figure2},
we then propose to introduce a mutual learning framework to jointly train \textsc{Tranx} and $\textsc{Tranx}'$.
Need to add that our framework is also suitable for joint training of more than two models.

In order to facilitate the understanding of our framework,
we also depict its training procedure in Algorithm \ref{conjugateGradient}.
Using this framework,
we continuously enhance these two models via mutual distillation,
where two one-to-one knowledge transfers in reverse directions are synchronously executed at each training step (Lines \ref{line7}-\ref{line15}).
During the one-to-one transfer procedure,
we expect each model to fully learn knowledge from not only its own training data but also the other.

To this end,
we require each considered model to fit the training data and match the action prediction distributions of the other model.
In addition to the conventional MLE loss on training data, we introduce one Kullback-Leibler (KL) loss to quantify the action prediction distribution divergence between the considered model and the other one.
Suppose $\theta$ and $\theta'$ to be the parameter sets of \textsc{Tranx} and $\textsc{Tranx}'$,
we define the following objective function to update their parameters:
\begin{align}
	\label{equation1}
	&J(D, \theta)=\sum_{(\mathbf{x,a})\in D}\{J_\textrm{MLE}(\mathbf{x,a};\theta) +
	\\& \frac{\lambda}{T}\cdot \sum_{n\in  \mathbf{z}}\text{KL}(p(a'_{t'(n)}|a'_{<t'(n)}, \mathbf{x};\theta') ||
	p(a_{t(n)}|a_{<t(n)},\mathbf{x};\theta))
	\},  \notag
\end{align}
\begin{align}
	\label{equation2}
	&J(D, \theta')=\sum_{(\mathbf{x,a'})\in D}\{J_\textrm{MLE}(\mathbf{x,a'};\theta') +
	\\& \frac{\lambda}{T}\cdot \sum_{n\in\mathbf{z}}\text{KL}(p(a_{t(n)}|a_{<t(n)},\mathbf{x};\theta)|| p(a'_{t'(n)}|a'_{<t'(n)},\mathbf{x};\theta'))
	\},  \notag
\end{align}
where $t(n)$ and $t'(n)$ denote the timesteps of the AST node $n$ during the pre-order and breadth-first traversals of the AST $\mathbf{z}$, respectively,
$\text{KL}(\cdot||\cdot)$ is the KL divergence, and $\lambda$ is the coefficient used to control the impacts of different losses.
Note that both our KL terms act on the AST node-level outputs of different models,
that is, $p(a'_{t'(n)}|a'_{<t'(n)}, \mathbf{x}; \theta')$ and $p(a_{t(n)}|a_{<t(n)},\mathbf{x}; \theta)$, where the AST node $n$ may correspond to different timesteps of our two models.
Back to Figure \ref{figure1},
the node $n_6$ corresponds to the 6-th and 3-rd timesteps of pre-order and breadth-first traversals, respectively.

We repeat the above-mentioned knowledge transfer process, until both models converge.
\begin{figure*}[th] \centering
	
	\small
	\subfigure[\textbf{DJANGO}]{
		\includegraphics[width=0.22\textwidth,trim=50 540 290 50,clip]{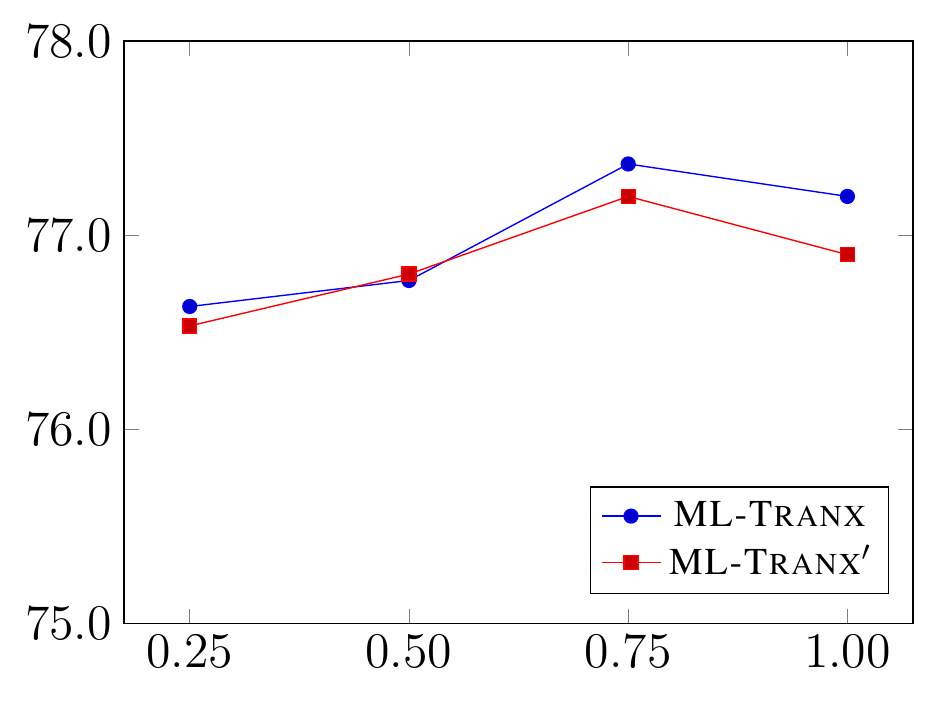}
	}
	~~~~
	\subfigure[\textbf{ATIS}]{
		\includegraphics[width=0.22\textwidth,trim=50 540 290 50,clip]{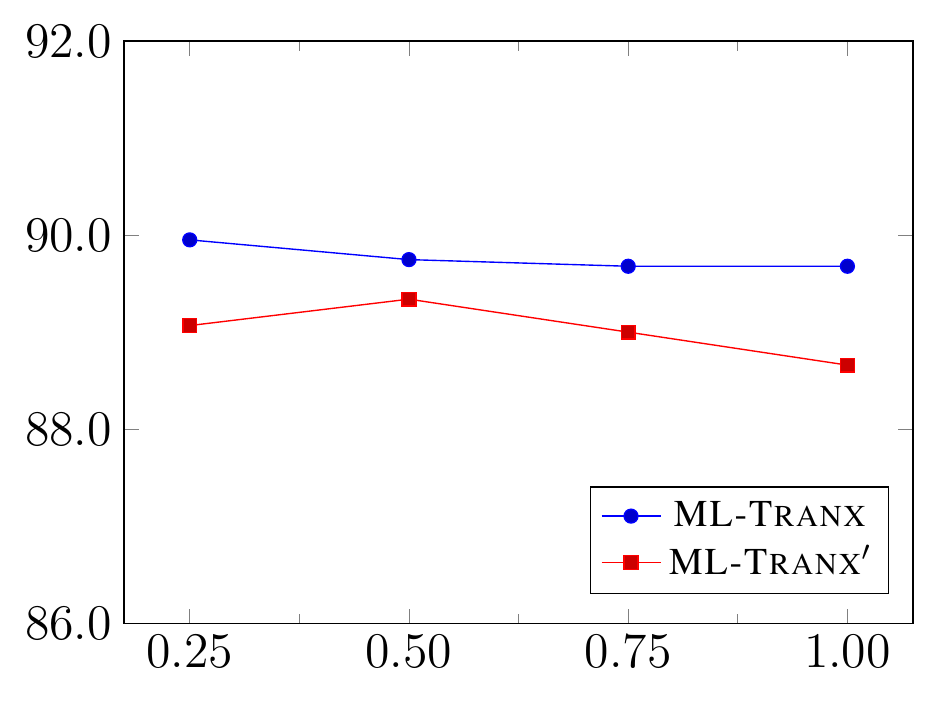}
	}
	~~~~
	\subfigure[\textbf{GEO}]{
		\includegraphics[width=0.22\textwidth,trim=50 540 290 50,clip]{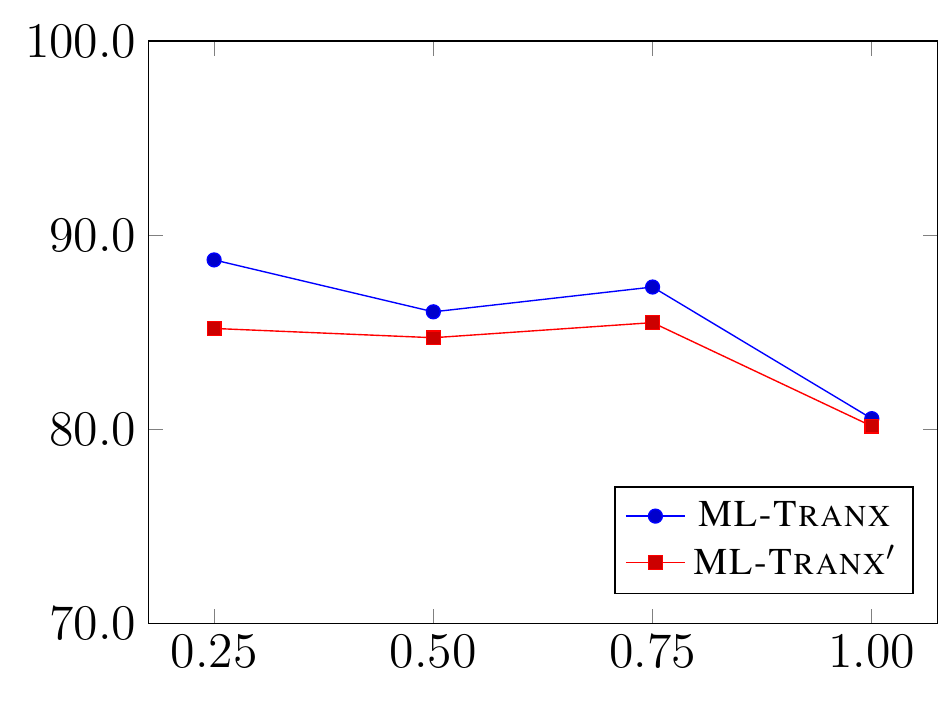}
	}
	~~~~
	\subfigure[\textbf{IFTTT}]{
		\includegraphics[width=0.22\textwidth,trim=50 540 290 50,clip]{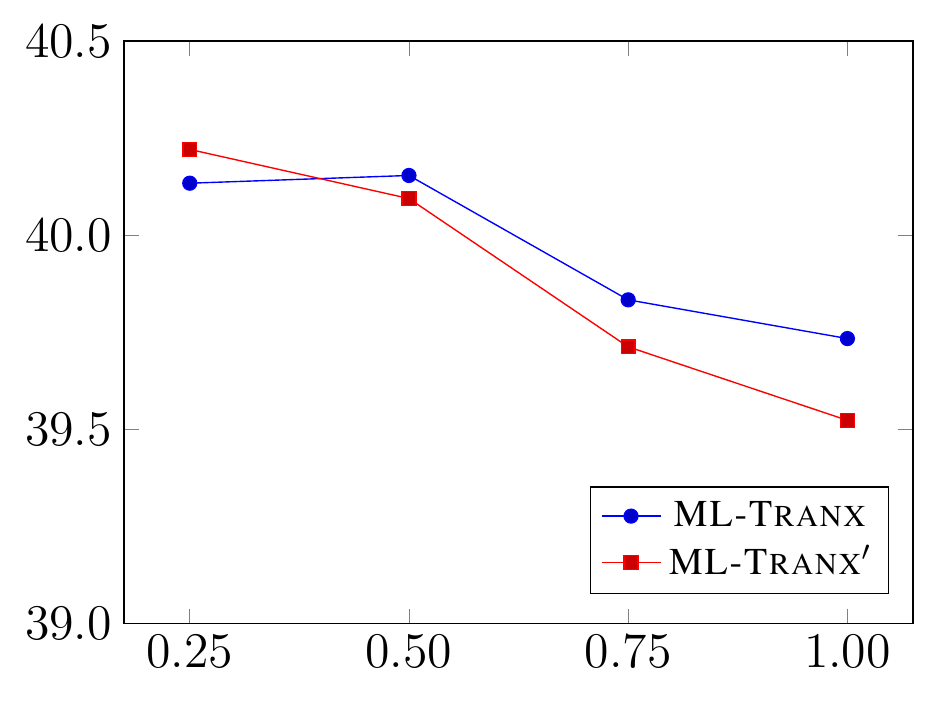}
	}
	\caption{Effects of $\lambda$ on the validation sets. }
	\label{figure3}
\end{figure*}

\section{Experiments}
\subsection{Datasets}

We carry out experiments on the following datasets: 1) \textbf{DJANGO} (Oda et al. 2015). This dataset consists of 18,805 lines of Python source code, with each line paired with an NL utterance.
We split the dataset into training/validation/test sets containing 16,000/1,000/1,805 instances, respectively.
2) \textbf{ATIS}. This dataset includes NL questions of a flight database, with each question is annotated with a lambda calculus query.
Following previous studies \cite{DBLP:conf/emnlp/YinN18,DBLP:conf/acl/YinN19,DBLP:conf/acl/XuJYVN20},
we use the standard splits of training/validation/test sets, which contains 4,473/491/448 instances, respectively.
3) \textbf{GEO}. It contains NL questions about US geography paired with corresponding Prolog database queries. we use the standard splits of 600/280 training/test instances.
4) \textbf{IFTTT} \cite{DBLP:conf/acl/QuirkMG15}.
It consists of if-this-then-that programs,
paired with NL utterances of their purpose.
The dataset is split into 68,083 training, 5,171 validation and 3,868 test instances.

\subsection{Baseline Models}
We apply mutual learning to simultaneously train \textsc{Tranx} and $\textsc{Tranx}'$, which are respectively referred to as \textsc{ML-Tranx} and  $\textsc{ML-Tranx}'$, avoiding the description confusion.
In addition to the conventional  \textsc{Tranx} and $\textsc{Tranx}'$,
and some commonly-used baselines (See Table \ref{performance}),
we compare the enhanced models with the following models:
\begin{itemize}

	\item \textbf{Ensemble} \cite{DBLP:conf/wmt/SennrichBCGHHBW17}. Using this method, we first employ depth-first and bread-first model independently to obtain two $k$-best lists, and then re-score the combination of these two lists from these two models.
	
	\item \textbf{MTL-Tranx} and $\textbf{MTL-Tranx}'$.  We apply multitask learning to jointly
	train \textsc{Tranx} and $\textsc{Tranx}'$ sharing the same encoder.
	
	\item \textbf{KD-\textsc{Tranx}}. We employ KD to transfer the knowledge of the $\textsc{Tranx}'$ with fixed parameters to enhance \textsc{Tranx}.
	
	\item \textbf{KD-$\textsc{Tranx}'$}. A $\textsc{Tranx}'$ enhanced by a fixed \textsc{Tranx} via KD.
	
	\item \textbf{ML2-\textsc{Tranx}}. Under our framework, two \textsc{Tranx} models with different initializations are
	jointly trained.
	
	\item \textbf{ML2-$\textsc{Tranx}'$}. Two $\textsc{Tranx}'$ models with different initializations are jointly trained under our framework.
\end{itemize}

Particularly, we report the average performance of two involved models for ML2-$\textsc{Tranx}$ and ML2-$\textsc{Tranx}'$.
We use the same experimental setup as \cite{DBLP:conf/acl/YinN17}. Specifically, we use 256 hidden units and 128-dimensional word vectors for NL utterance encoding,
and tune the dimension of various embeddings on validation datasets for each corpus.
We initialize all parameters by uniformly sampling within the interval [-0.1, 0.1]. Besides, we set the batch size as 10 and employ dropout after each layer,
where the drop rate is sequentially set to 0.5, 0.3, 0.4, and 0.3 for our four datasets, respectively.
To alleviate the instability of the model training, we run each model five times and report the average performance
in terms of exact matching accuracy.
Following the evaluation protocol in \cite{DBLP:conf/acl/BeltagyQ16} for accuracies, we evaluate model performance on IFTTT at both channel and full parse tree (channel + function) levels.

\subsection{Effect of $\lambda$}

As shown in Equations \ref{equation1} and \ref{equation2},
the coefficient $\lambda$ is an important hyper-parameter that controls relative impacts of the
MLE loss and the divergence losses.
Thus,
we first investigate the effect of $\lambda$ on our proposed framework.
To this end,
we gradually vary $\lambda$ from 0 to 1 with an increment of 0.25 at each step,
and report the performance of our models using different $\lambda$s on the validation datasets.
Since there exists no validation set in GEO,
we temporarily split the training data into two parts: 480 instances for training and 120 instances for validation, and use them to determine the optimal $\lambda$.
However,
with the optimal $\lambda$,
we still use the whole training data in subsequent experiments.

Figure \ref{figure3} shows the experimental results.
According to the average performance of our models on four validation sets,
we set $\lambda$ as 0.75, 0.5, 0.25 and 0.25 for our four datasets in all experiments thereafter.

\subsection{Main Results}

\begin{table*}[ht]
	\centering

		\begin{threeparttable}[width=1\textwidth]
			\scalebox{0.85}{	
			\begin{tabular}{lcccccclll}
				\toprule
				\multicolumn{1}{c}{\multirow{2}{*}{\textbf{Model}}}&{\bf DJANGO}&{\bf ATIS}&{\bf GEO}&{\bf IFTTT}\cr
				\cmidrule(r){2-6}
				&\textsc{Acc.} &\textsc{Acc.}&\textsc{Acc.}& \textsc{Acc. CHANNEL / FULL TREE}  \cr
				\midrule
				\textsc{Drnn} \cite{DBLP:conf/iclr/Alvarez-MelisJ17} &--&--&--& 90.1 / 78.2  \cr
				\textsc{Asn} \cite{DBLP:conf/acl/RabinovichSK17} &--& 85.9&87.1& -- / -- &\cr
				\textsc{Seq2AST-YN17} \cite{DBLP:conf/acl/YinN17} & 75.8 & -- & -- & 90.0 / 82.0 \cr
				\textsc{Coarse2Fine} \cite{DBLP:conf/acl/LapataD18} & 74.1 &{87.7}&{88.2}& -- / -- \cr
				\textsc{Tranx} \cite{DBLP:conf/acl/YinN19} & 77.3{\tiny$\pm$0.4}&87.6{\tiny$ \pm$0.1}&88.8{\tiny$\pm$1.0}& -- / -- \cr
				\textsc{TreeConv} \cite{DBLP:conf/aaai/SunZMXLZ19} &--&85.0&--& -- / -- \cr
				\textsc{TreeGen} \cite{DBLP:conf/aaai/SunZXSMZ20} &--&{89.1}&\textbf{{89.6}}& -- / -- \cr
				\midrule
				\textsc{Ensemble} \cite{DBLP:conf/wmt/SennrichBCGHHBW17} &78.6{\tiny$ \pm$0.4}&88.3{\tiny$\pm$0.5}&88.8{\tiny$\pm$0.7}& 91.1{\tiny$\pm$1.0} / 84.5{\tiny$\pm$0.5}\cr
				\midrule
				\textsc{Tranx} & 77.3{\tiny$ \pm$0.4}&87.7{\tiny$\pm$0.5}&88.7{\tiny$\pm$0.7}& 91.0{\tiny$\pm$1.0} / 82.8{\tiny$\pm$0.5}\cr
				$\textsc{MTL-Tranx}$ &78.2{\tiny$\pm$0.1}&88.0{\tiny$\pm$0.3}&89.0{\tiny$\pm$0.8}& 91.6{\tiny$\pm$0.2} / 84.3{\tiny$\pm$1.4}\cr
				$\textsc{KD-Tranx}$ &78.1{\tiny$\pm$0.3}&87.9{\tiny$\pm$0.3}&88.8{\tiny$\pm$0.6}& 91.2{\tiny$\pm$0.5} / 83.8{\tiny$\pm$0.5}\cr
				$\textsc{ML2-Tranx}$ &79.0{\tiny$\pm$0.1}&87.8{\tiny$\pm$1.1}&88.9{\tiny$\pm$0.6}& 91.5{\tiny$\pm$0.4} / 84.9{\tiny$\pm$0.1}\cr
				$\textsc{ML-Tranx}$ &\textbf{79.6}{\tiny$\pm$0.3}&\textbf{89.3}{\tiny$\pm$0.3}&89.2{\tiny$\pm$0.6}& \textbf{92.0}{\tiny$\pm$0.3} / 85.2{\tiny$\pm$1.6}  \cr		
				\midrule
				$\textsc{Tranx}'$ &76.8{\tiny$\pm$0.2}&86.9{\tiny$\pm$0.3}&87.0{\tiny$\pm$1.3}& 90.1{\tiny$\pm$0.2} / 80.1{\tiny$\pm$1.0} \cr
				$\textsc{MTL-Tranx}'$&78.0{\tiny$\pm$0.1}&87.7{\tiny$\pm$0.1}&88.6{\tiny$\pm$0.5}& 91.7{\tiny$\pm$0.1} / 84.7{\tiny$\pm$0.7}\cr		
				$\textsc{KD-Tranx}'$ &77.6{\tiny$\pm$0.4}&87.1{\tiny$\pm$0.4}&88.3{\tiny$\pm$0.4}& 90.8{\tiny$\pm$0.3} / 83.4{\tiny$\pm$0.8}\cr		
				$\textsc{ML2-Tranx}'$ &77.8{\tiny$\pm$0.2}&87.0{\tiny$\pm$1.3}&87.9{\tiny$\pm$0.3}& 91.0{\tiny$\pm$0.3} / 83.8{\tiny$\pm$1.0}\cr	
				$\textsc{ML-Tranx}'$ &78.6{\tiny$\pm$0.1}&88.4{\tiny$ \pm$0.2}&88.9{\tiny$\pm$0.4}& 91.9{\tiny$\pm$0.1} / \textbf{85.7}{\tiny$\pm$0.3}\cr
				\bottomrule
			\end{tabular}
		}
		\end{threeparttable}
		\caption{Main experimental results. All results shown in the upper part are directly cited from their corresponding papers.}
		\label{performance}

\end{table*}

Table \ref{performance} reports the overall experimental results.
Both models involved are improved under our framework.
Besides,
we can draw the following conclusions:

(1) Our implemented $\textsc{Tranx}$ achieves comparable performance to \cite{DBLP:conf/acl/YinN19}.
Thus, we confirm that our reimplemented baseline is competitive.

(2) Both $\textsc{ML-Tranx}$ and $\textsc{ML-Tranx}'$ significantly surpass their respective important baselines: \textsc{Tranx} and $\textsc{Tranx}'$.
Besides,
both of our enhanced models surpass \textsc{Drnn}.
These results strongly validate the effectiveness of our proposed training framework.

(3) Both $\textsc{ML-Tranx}$ and $\textsc{ML-Tranx}'$ outperforms Ensemble. This result is reasonable since $\textsc{Tranx}$ and $\textsc{Tranx}'$ can interact with each other via mutual learning while they are independently trained when using ensemble.

(4) Compared with multitask learning based models: $\textsc{MTL-Tranx}$ and $\textsc{MTL-Tranx}'$, both $\textsc{ML-Tranx}$ and $\textsc{ML-Tranx}'$ achieve better performance. This is due to the fact that in addition to sharing encoder parameters, mutual learning can refine model training by working as regularization terms, which has also been mentioned in \cite{DBLP:journals/corr/RomeroBKCGB14,DBLP:conf/acl/ClarkLKML19}.

(5) $\textsc{ML-Tranx}$ and $\textsc{ML-Tranx}'$ respectively show better performance than $\textsc{KD-Tranx}$ and $\textsc{KD-Tranx}'$, verifying the necessity of multiple iterations of knowledge transfer.

(6) $\textsc{ML-Tranx}$ and $\textsc{ML-Tranx}'$ perform better in different datasets. The underlying reason is that each instance contains only three layers in IFTTT, where the effectiveness of pre-order decoding is greatly restricted.
\begin{table}[H]
	\centering
	
		\begin{threeparttable}[width=0.5\textwidth]	
			\scalebox{0.85}{
			\begin{tabular}{lcccccclll}	
				\toprule
				\multicolumn{1}{c}{\multirow{2}{*}{\textbf{Model}} } &\multicolumn{5}{c}{\bf DJANGO}\cr
				\cmidrule(r){2-6} & $[1$, $10]$  &  $[11$, $20]$ &  $[21$, $30]$  & $[31$, $40]$  & $[41$, $\infty)$   \cr
				\midrule
				\textsc{Ensemble} & 92.9 & 79.7 & 42.9 &\textbf{ 10.9} & -- \cr
				\midrule
				\textsc{Tranx} & 93.0 & 76.7 & 45.7 & 5.4 & -- \cr
				$\textsc{MTL-Tranx}$ & 94.7 & 81.5 & 44.0 & 8.7 & -- \cr
				$\textsc{KD-Tranx}$ & 93.8 & 79.1 & 44.0 & 6.5 & -- \cr
				$\textsc{ML2-Tranx}$ &93.8  &  80.9 &  44.3 &  6.5 & --\cr
				$\textsc{ML-Tranx}$ &\textbf{95.0}  & \textbf{82.2} &  \textbf{48.6}  & 8.7 & --\cr	
				\midrule
				$\textsc{Tranx}'$ &92.4 & 79.1 &  42.3 & \textbf{ 10.9} & --\cr
				$\textsc{MTL-Tranx}'$ & 94.8 & 81.1 & 44.6 & 6.5 & -- \cr	
				$\textsc{KD-Tranx}'$ & 93.8 & 80.9 & 43.4 & 8.7 & -- \cr
				$\textsc{ML2-Tranx}'$ &93.8 &80.0 & 46.0 & \textbf{10.9}&--\cr
				$\textsc{ML-Tranx}'$&94.0 & 81.8 & 43.4 & \textbf{10.9}&--\cr
				\bottomrule
			\end{tabular}}
		\end{threeparttable}
		\caption{Accuracy on different groups of DJANGO.
			Please note that on the group [41, $\infty)$, none of the models performed well.}
		\label{2}
	
\end{table}
(7) Compared with $\textsc{ML2-Tranx}$,
our \textsc{Tranx} still exhibits better performance.
Likewise,
$\textsc{Tranx}$ performs better than $\textsc{ML2-Tranx}'$.
Hence,
we confirm that $\textsc{Tranx}$ and $\textsc{Tranx}'$ are indeed able to benefit each other.

\subsection{Performance by the size of AST}
Further,
in order to inspect the generality of our proposed framework,
we follow Yin and Neubig \shortcite{DBLP:conf/acl/YinN17} to split datasets into different groups according to their AST sizes,
and then compare the model performance at each group.

Tables \ref{2}, \ref{3} and \ref{4} display the results on DJANGO, ATIS and GEO, respectively. \footnote{We do not display the results on IFTTT, since all of its instances have the same AST size.}
\begin{table}[H]
	\centering
	
		\begin{threeparttable}[width=0.5\textwidth]
				\scalebox{0.85}{
			\begin{tabular}{lcccccclll}	
				\toprule
				\multicolumn{1}{c}{\multirow{2}{*}{\textbf{Model}} }&\multicolumn{5}{c}{\bf ATIS}\cr
				\cmidrule(r){2-6} & $[1$, $10]$  &  $[11$, $20]$ &  $[21$, $30]$  & $[31$, $40]$  & $[41$, $\infty)$   \cr
				\midrule
				\textsc{Ensemble} & \textbf{93.3} & 78.9 & \textbf{96.8} & 93.6 & 60.9 \cr
				\midrule
				\textsc{Tranx} & 88.3 & 84.2 & 95.3 & 92.8 & 60.9 \cr
				$\textsc{MTL-Tranx}$ & \textbf{93.3} & 89.5 & 94.5 & 92.8 & 65.2 \cr
				$\textsc{KD-Tranx}$ & 90.8 & 84.2 & 95.3 & 93.6 & 60.9 \cr
				$\textsc{ML2-Tranx}$ &90.0 & 88.0 & 96.0 & 93.6 & 60.9\cr
				$\textsc{ML-Tranx}$ &\textbf{93.3} &\textbf{ 92.1} & \textbf{96.8} &  93.6 & \textbf{63.8}\cr
				\midrule
				$\textsc{Tranx}'$ &85.0 & 84.2 & 96.0 & 92.8 & 56.5\cr
				$\textsc{MTL-Tranx}'$ & 90.0 & 89.5 & 95.3 & \textbf{94.1} & 56.5 \cr
				$\textsc{KD-Tranx}'$ & 90.0 & 88.2 &96.0 & 92.8 & 60.1 \cr
				$\textsc{ML2-Tranx}'$ &89.2 & 88.2 & 96.0 & 92.2 & 60.1\cr
				$\textsc{ML-Tranx}'$&91.7 & 89.5 & 96.0 & 93.6 & 60.9\cr
				\bottomrule
			\end{tabular}}
		\end{threeparttable}
		\caption{Accuracy on different groups of ATIS.}
		\label{3}

\end{table}
\begin{table}[H]
	\centering
	
		\begin{threeparttable}[width=0.1\textwidth]
			\scalebox{0.85}{	
			\begin{tabular}{lcccccclll}
				\toprule
				\multicolumn{1}{c}{\multirow{2}{*}{\textbf{Model}}}&\multicolumn{5}{c}{\bf GEO} \\
				\cmidrule(r){2-6}&$[1$, $10]$&$[11$, $20]$&$[21$, $30]$&$[31$, $40]$&$[41$, $\infty)$\\
				\midrule
				\textsc{Ensemble}&\textbf{98.1}&\textbf{93.8}&90.1&77.1&45.5\cr
				\midrule
				\textsc{Tranx}&\textbf{98.1}&\textbf{93.8}&88.7&72.2&45.5\cr
				$\textsc{MTL-Tranx}$&96.3&93.8&92.9&80.6&\textbf{54.5} \cr
				$\textsc{KD-Tranx}$&97.2&93.8&92.9&80.6& \textbf{54.5}\cr
				$\textsc{ML2-Tranx}$&\textbf{98.1}&\textbf{93.8}&92.9&80.6 & 45.5\cr
				$\textsc{ML-Tranx}$&\textbf{98.1}&92.6& \textbf{93.9}& \textbf{83.3} & \textbf{54.5}\cr
				\midrule
				$\textsc{Tranx}'$&96.3&92.6&88.8& 72.2  & 45.5\cr
				$\textsc{MTL-Tranx}'$&96.3&92.6& 89.8 & 80.6 & 36.4 \cr
				$\textsc{KD-Tranx}'$&96.3&93.2& 88.8  & 72.2 & 40.9 \cr
				$\textsc{ML2-Tranx}'$&97.2& 93.2& 90.1 & 80.6 & 40.9\cr
				$\textsc{ML-Tranx}'$&\textbf{98.1}&\textbf{93.8} & 91.8 & 80.6 &  45.5\cr
				\bottomrule
			\end{tabular}
		}
		\end{threeparttable}
		\caption{Accuracy on different groups of GEO.}
		\label{4}

\end{table}

\begin{table}[ht] 
	\centering

	\begin{threeparttable}
		\scalebox{0.90}{	
		\begin{tabular}{cp{6.5cm}} 
			\toprule
			\multirow{2.5}{*}{Source}  & {call the method line.lstrip [ line . lstrip ] , if the result starts with TRANSLATOR\_COMMENT\_MARK , }\\
			\hline
			\midrule 
			\multirow{2}{*}{$\textsc{Tranx}$} & {if \textcolor{red}{not line.lstrip}(TRANSLATOR\_COMMENT\_ MARK ): pass \ \ \ }\textcolor{red}{\xmark}
			\\
			\midrule
			\multirow{2.8}{*}{$\textsc{Tranx}'$}  & {if line.lstrip(\textcolor{red}{TRANSLATOR\_COMMENT\_MA RK}).startswith(TRANSLATOR\_COMMENT\_ MARK): pass \ \ \ }\textcolor{red}{\xmark}
			\\	
			\midrule
			\multirow{2}{*}{$\textsc{Ensemble}$}  & {if \textcolor{red}{line.}startswith(TRANSLATOR\_COMMENT \_MARK): pass \ \ \ }\textcolor{red}{\xmark}
			\\
			\midrule
			\multirow{2}{*}{$\textsc{MTL-Tranx}$}  & {if line.lstrip().\textcolor{red}{starts}(TRANSLATOR\_COMME NT\_MARK):pass \ \ \ }\textcolor{red}{\xmark}
			\\	
			\midrule
			\multirow{2}{*}{$\textsc{KD-Tranx}$}  & {if \textcolor{red}{line.}startswith(TRANSLATOR\_COMMENT \_MARK): pass \ \ \ }\textcolor{red}{\xmark}
			\\
			\midrule
			\multirow{2}{*}{$\textsc{ML2-Tranx}$}  &{if \textcolor{red}{not line.}startswith(TRANSLATOR\_COMME NT\_MARK): pass \ \ \ }\textcolor{red}{\xmark}
			\\
			\midrule
			\multirow{2}{*}{$\textsc{ML-Tranx}$}  &{if \textcolor{blue}{line.lstrip().startswith}(TRANSLATOR\_COM MENT\_MARK): pass \ \ \ }\textcolor{blue}{\cmark}
			\\
			\bottomrule
		\end{tabular}
	}
	\end{threeparttable}
	\caption{Codes produced by different models.
		Incorrect codes are marked in red while counterparts are marked in blue. }
	\label{usecase}
\end{table}

We can observe that our enhanced models outperform or achieve comparable performance than their corresponding baselines respectively on almost all groups of datasets.
Thus,
we confirm again the effectiveness and generality of our proposed framework.

\subsection{Case Study}

We compare the $1$-best codes produced by different code generation models, so as to better understand how our model outperforms others.
Table \ref{usecase} shows an example of codes produced from Django dataset.
We find that the model based on depth-first traversal tends to generate the wrong function names,
while $\textsc{Tranx}'$ prefers to makes a mistake on the generation of argument.
Although using various strategies,
most of models are unable to generate completely correct code.
By contrast,
our model $\textsc{ML-Tranx}$ absorbs advantages of models with decoders based on different traversals and succeed in overcoming all these problems to generate more accurate code.

\section{Related Work}
With the rapid development of deep learning, neural network based models have now become dominant in code generation.
In this aspect, Ling et al., \shortcite{DBLP:conf/acl/LingBGHKWS16} considered code generation as a conditional text generation task, which can be solved by a neural sequence-to-sequence model. To exploit syntactic and semantic constraints of the programs, both of which are important for code generation,
more researchers resorted to neural Seq2Tree models transforming each NL utterance into a sequence of AST based grammar actions.
For example,
Yin and Neubig \shortcite{DBLP:conf/acl/YinN17,DBLP:conf/emnlp/YinN18} proposed Seq2Tree models to generate tree-structured MRs using a series of tree-construction actions.
Then, Sun et al. \shortcite{DBLP:conf/aaai/SunZMXLZ19,DBLP:conf/aaai/SunZXSMZ20} introduced CNN network and Transformer architecture to handle the long dependency problem.
From a different perspective, Yin and Neubig \shortcite{DBLP:conf/acl/YinN19} explored reranking an $N$-best list of candidate results for adequacy and coherence of the programs.
Meanwhile,
other approaches have been extensively investigated.
Wei et al., \shortcite{DBLP:conf/nips/WeiL0FJ19} applied dual training to jointly model code summarization and code generation,
where the specific intuitive correlation between these two tasks can be fully leveraged to benefit each other.

Significantly different from the above work,
we further explore the Seq2Tree model with breadth-first traversal based decoding.
More importantly,
we introduce mutual learning based model training framework to iteratively enhance multiple models via knowledge distillation.
Mutual learning has been successfully used in image recognition \cite{DBLP:conf/cvpr/ZhangXHL18,DBLP:conf/icml/FurlanelloLTIA18,DBLP:conf/nips/LanZG18},
machine translation \cite{Zeng:EMNLP2019},
machine reading comprehension \cite{Liu:IJCAI2020},
and sentiment analysis \cite{DBLP:conf/aaai/XueZZ20}.
However, to the best of our knowledge, our work is the first attempt to explore this approach for code generation.
Among all models for code generation,
the most related to our is \cite{DBLP:conf/iclr/Alvarez-MelisJ17}.
This work proposed to use doubly-recurrent decoders (vertical and horizontal LSTMs) trained on ASTs,
of which outputs can be concatenated to exploit context in different dimensions.
Compared with this work, ours still significantly differs from it in following aspects:
1) Our approach only affects the model training while has no impact on the model testing;
2) During the model testing, David Alvarez{-}Melis and Jaakkola. \shortcite{DBLP:conf/iclr/Alvarez-MelisJ17} simultaneously use two LSTMs with more parameters while we only use one LSTM decoder,
and
3) Experimental results show both of our enhanced models are able to surpass \cite{DBLP:conf/iclr/Alvarez-MelisJ17}.

\section{Conclusion and Future Work}

In this paper, we have explored the Seq2Tree model with breadth-first traversal based decoding and proposed a mutual learning based model training framework for code generation, where the models with different traversals based decodings can be enhanced simultaneously.
Experimental results and in-depth analyses on several commonly-used datasets strongly demonstrate the effectiveness and generality of our proposed framework.

In the future, we plan to refine our framework via self-distillation \cite{wei-etal-2019-online}.
Moreover,
we will explore asynchronous bidirectional decoding \cite{Zhang:AAAI2018,Su:AI2019}to combine advantages of code generation models with different traversals based decodings (preorder traversal vs breadth-first traversal).

\section{Acknowledgements}
The project was supported by
National Natural Science Foundation of China (No. 62036004, No. 61672440),
Natural Science Foundation of Fujian Province of China (No.2020J06001)
and College-level Undergraduate Innovation Training Project of School of Informatics Xiamen University (No.2020Y1071).

\bibliography{mysef}

\end{document}